# On the Sample Complexity of Reinforcement Learning with a Generative Model

**Mohammad Gheshlaghi Azar**  M.AZAR@SCIENCE.RU.NL
Department of Biophysics, Radboud University Nijmegen, 6525 EZ Nijmegen, The Netherlands

**Rémi Munos**  REMI.MUNOS@INRIA.FR
INRIA Lille, SequeL Project, 40 avenue, Halley 59650, Villeneuve dAscq, France

**Hilbert J. Kappen**  B.KAPPEN@SCIENCE.RU.NL
Department of Biophysics, Radboud University Nijmegen, 6525 EZ Nijmegen, The Netherlands

## Abstract

We consider the problem of learning the optimal action-value function in the discounted-reward Markov decision processes (MDPs). We prove a new PAC bound on the sample-complexity of model-based value iteration algorithm in the presence of the generative model, which indicates that for an MDP with $N$ state-action pairs and the discount factor $\gamma \in [0,1)$ only $O\big(N \log(N/\delta)/\big((1-\gamma)^3 \varepsilon^2\big)\big)$ samples are required to find an $\varepsilon$-optimal estimation of the action-value function with the probability $1-\delta$. We also prove a matching lower bound of $\Theta\big(N\log(N/\delta)/\big((1-\gamma)^3\varepsilon^2\big)\big)$ on the sample complexity of estimating the optimal action-value function by every RL algorithm. To the best of our knowledge, this is the first matching result on the sample complexity of estimating the optimal (action-) value function in which the upper bound matches the lower bound of RL in terms of $N$, $\varepsilon$, $\delta$ and $1/(1-\gamma)$. Also, both our lower bound and our upper bound significantly improve on the state-of-the-art in terms of $1/(1-\gamma)$.

## 1. Introduction

Model-based value iteration (VI) (Kearns & Singh, 1999; Buş,oniu et al., 2010) is a well-known reinforcement learning (RL) (Szepesvári, 2010; Sutton & Barto, 1998) algorithm which relies on an empirical estimate of the state-transition distributions to estimate the optimal (action-)value function through the Bellman recursion. In the finite state-action problems, it has been shown that an action-value based variant of VI, model-based Q-value iteration (QVI), finds an $\varepsilon$-optimal estimate of the action-value function with high probability using only $T = \tilde{O}(N/\big((1-\gamma)^4 \varepsilon^2\big))$ samples (Kearns & Singh, 1999; Kakade, 2004, chap. 9.1), where $N$ and $\gamma$ denote the size of state-action space and the discount factor, respectively.[1] Although this bound matches the best existing upper bound on the sample complexity of estimating the action-value function (Azar et al., 2011a), it has not been clear, so far, whether this bound is a tight bound on the performance of QVI or it can be improved by a more careful analysis of QVI algorithm. This is mainly due to the fact that there is a gap of order $1/(1-\gamma)^2$ between the upper bound of QVI and the state-of-the-art result for lower bound, which is of $\Omega\big(N/\big((1-\gamma)^2\varepsilon^2\big)\big)$ (Azar et al., 2011b).

In this paper, we focus on the problems which are formulated as finite state-action discounted infinite-horizon Markov decision processes (MDPs), and prove a new tight bound of $O\big(N\log(N/\delta)/\big((1-\gamma)^3\varepsilon^2\big)\big)$ on the sample complexity of the QVI algorithm. The new upper bound improves on the existing bound of QVI by an order of $1/(1-\gamma)$.[2] We also present a new matching lower bound of $\Theta\big(N\log(N/\delta)/\big((1-\gamma)^3\varepsilon^2\big)\big)$, which also improves on the best existing lower bound of RL by an order of $1/(1-\gamma)$. The new results, which close the above-mentioned gap between the lower bound and the upper bound of RL, guarantee that no learning method, given a generative model of the MDP, can be significantly more efficient than QVI in terms of the sample complexity of estimating the action-value function.

---

[1] The notation $g = \tilde{O}(f)$ implies that there are constants $c_1$ and $c_2$ such that $g \leq c_1 f \log^{c_2}(f)$.

[2] In this paper, to keep the presentation succinct, we only consider the value iteration algorithm. However, one can prove upper-bounds of a same order for other model-based methods such as policy iteration and linear programming using the results of this paper.

The main idea to improve the upper bound of QVI is to express the performance loss of QVI in terms of the variance of the sum of discounted rewards as opposed to the maximum $V_{\max} = R_{\max}/(1-\gamma)$ in the previous results. For this we make use of Bernstein's concentration inequality (Cesa-Bianchi & Lugosi, 2006, appendix, pg. 361), which bounds the estimation error in terms of the variance of the value function. We also rely on the fact that the variance of the sum of discounted rewards, like the expected value of the sum (value function), satisfies a Bellman-like equation, in which the variance of the value function plays the role of the instant reward (Munos & Moore, 1999; Sobel, 1982), to derive a sharp bound on the variance of the value function. In the case of lower bound, we improve on the result of Azar et al. (2011b) by adding some structure to the class of MDPs for which we prove the lower bound: In the new model, there is a high probability for transition from every intermediate state to itself. This adds to the difficulty of estimating the value function, since even a small estimation error may propagate throughout the recursive structure of the MDP and inflict a big performance loss especially for $\gamma$'s close to 1.

The rest of the paper is organized as follows. After introducing the notation used in the paper in Section 2, we describe the *model-based Q-value iteration* (QVI) algorithm in Subsection 2.1. We then state our main theoretical results, which are in the form of PAC sample complexity bounds in Section 3. Section 4 contains the detailed proofs of the results of Sections 3, i.e., sample complexity bound of QVI and a general new lower bound for RL. Finally, we conclude the paper and propose some directions for the future work in Section 5.

## 2. Background

In this section, we review some standard concepts and definitions from the theory of Markov decision processes (MDPs). We then present the model-based Q-value iteration algorithm of Kearns & Singh (1999). We consider the standard reinforcement learning (RL) framework (Bertsekas & Tsitsiklis, 1996; Sutton & Barto, 1998) in which a learning agent interacts with a stochastic environment and this interaction is modeled as a discrete-time discounted MDP. A dis-counted MDP is a quintuple $(\mathcal{X}, \mathcal{A}, P, \mathcal{R}, \gamma)$, where $\mathcal{X}$ and $\mathcal{A}$ are the set of states and actions, $P$ is the state transition distribution, $\mathcal{R}$ is the reward function, and $\gamma \in (0,1)$ is a discount factor. We denote by $P(\cdot|x,a)$ and $r(x,a)$ the probability distribution over the next state and the immediate reward of taking action $a$ at state $x$, respectively.

**Remark 1.** *To keep the representation succinct, in the sequel, we use the notation $\mathcal{Z}$ for the joint state-action space $\mathcal{X} \times \mathcal{A}$. We also make use of the shorthand notations $z$ and $\beta$ for the state-action pair $(x,a)$ and $1/(1-\gamma)$, respectively.*

**Assumption 1** (MDP Regularity). *We assume $\mathcal{Z}$ and, subsequently, $\mathcal{X}$ and $\mathcal{A}$ are finite sets with cardinalities $N$, $|\mathcal{X}|$ and $|\mathcal{A}|$, respectively. We also assume that the immediate reward $r(x,a)$ is taken from the interval $[0,1]$.*

A mapping $\pi : \mathcal{X} \to \mathcal{A}$ is called a stationary and deterministic Markovian policy, or just a policy in short. Following a policy $\pi$ in an MDP means that at each time step $t$ the control action $A_t \in \mathcal{A}$ is given by $A_t = \pi(X_t)$, where $X_t \in \mathcal{X}$. The *value* and the *action-value functions* of a policy $\pi$, denoted respectively by $V^\pi : \mathcal{X} \to \mathbb{R}$ and $Q^\pi : \mathcal{Z} \to \mathbb{R}$, are defined as the expected sum of discounted rewards that are encountered when the policy $\pi$ is executed. Given an MDP, the goal is to find a policy that attains the best possible values, $V^*(x) \triangleq \sup_\pi V^\pi(x)$, $\forall x \in \mathcal{X}$. Function $V^*$ is called the *optimal value function*. Similarly the *optimal action-value function* is defined as $Q^*(x,a) = \sup_\pi Q^\pi(x,a)$. We say that a policy $\pi^*$ is optimal if it attains the optimal $V^*(x)$ for all $x \in \mathcal{X}$. The policy $\pi$ defines the state transition kernel $P_\pi$ as: $P_\pi(y|x) \triangleq P(y|x,\pi(x))$ for all $x \in \mathcal{X}$. The right-linear operators $P^\pi\cdot$, $P\cdot$ and $P_\pi\cdot$ are then defined as $(P^\pi Q)(z) \triangleq \sum_{y \in \mathcal{X}} P(y|z) Q(y,\pi(y))$, $(PV)(z) \triangleq \sum_{y \in \mathcal{X}} P(y|z) V(y)$ for all $z \in \mathcal{Z}$ and $(P_\pi V)(x) \triangleq \sum_{y \in \mathcal{X}} P_\pi(y|x) V(y)$ for all $x \in \mathcal{X}$, respectively. Finally, $\|\cdot\|$ shall denote the supremum ($\ell_\infty$) norm which is defined as $\|g\| \triangleq \max_{y \in \mathcal{Y}} |g(y)|$, where $\mathcal{Y}$ is a finite set and $g : \mathcal{Y} \to \mathbb{R}$ is a real-valued function.[3]

### 2.1. Model-based Q-value Iteration (QVI)

The algorithm makes $n$ transition samples from each state-action pair $z \in \mathcal{Z}$ for which it makes $n$ calls to the generative model.[4] It then builds an empirical model of the transition probabilities as: $\widehat{P}(y|z) \triangleq m(y,z)/n$,

---

[3]For ease of exposition, in the sequel, we remove the dependence on $z$ and $x$, e.g., writing $Q$ for $Q(z)$ and $V$ for $V(x)$, when there is no possible confusion.

[4]The total number of calls to the generative model is given by $T = nN$.

where $m(y, z)$ denotes the number of times that the state $y \in \mathcal{X}$ has been reached from $z \in \mathcal{Z}$. The algorithm then makes an empirical estimate of the optimal action-value function $Q^*$ by iterating some action-value function $Q_k$, with the initial value of $Q_0$, through the empirical Bellman optimality operator $\widehat{\mathcal{T}}$.[5]

## 3. Main Results

Our main results are in the form of PAC (probably approximately correct) bounds on the $\ell_\infty$-norm of the difference of the optimal action-value function $Q^*$ and its sample estimate:

**Theorem 1** (PAC-bound for model-based Q-value iteration). *Let Assumption 1 hold and $T$ be a positive integer. Then, there exist some constants $c$ and $c_0$ such that for all $\varepsilon \in (0, 1)$ and $\delta \in (0, 1)$ a total sampling budget of*
$$T = \lceil \frac{c\beta^3 N}{\varepsilon^2} \log \frac{c_0 N}{\delta} \rceil,$$
*suffices for the uniform approximation error $\|Q^* - Q_k\| \leq \varepsilon$, w.p. (with the probability) at least $1 - \delta$, after only $k = \lceil \log(6\beta/\varepsilon)/\log(1/\gamma) \rceil$ iteration of QVI algorithm.[6] In particular, one may choose $c = 68$ and $c_0 = 12$.*

The following general result provides a tight lower bound on the number of transitions $T$ for every RL algorithm to achieve an $\varepsilon$-optimal estimate of the action-value function w.p. $1 - \delta$, under the assumption that the algorithm is $(\varepsilon, \delta, T)$-correct:

**Definition 1** ($(\varepsilon, \delta, T)$-correct algorithm). *Let $Q_T^\mathfrak{A}$ be the estimate of $Q^*$ by an RL algorithm $\mathfrak{A}$ after observing $T \geq 0$ transition samples. We say that $\mathfrak{A}$ is $(\varepsilon, \delta, T)$-correct on the class of MDPs $\mathbb{M}$ if $\|Q^* - Q_T^\mathfrak{A}\| \leq \varepsilon$ with probability at least $1 - \delta$ for all $M \in \mathbb{M}$.[7]*

**Theorem 2** (Lower bound on the sample complexity of estimating the optimal action-value function). *There exists some constants $\varepsilon_0$, $\delta_0$, $c_1$, $c_2$, and a class of MDPs $\mathbb{M}$, such that for all $\varepsilon \in (0, \varepsilon_0)$, $\delta \in (0, \delta_0/N)$, and every $(\varepsilon, \delta, T)$-correct RL algorithm $\mathfrak{A}$ on the class of MDPs $\mathbb{M}$ the number of transitions needs to be at least*
$$T = \lceil \frac{\beta^3 N}{c_1 \varepsilon^2} \log \frac{N}{c_2 \delta} \rceil.$$

---

[5]The operator $\widehat{\mathcal{T}}$ is defined on the action-value function $Q$, for all $z \in \mathcal{Z}$, by $\widehat{\mathcal{T}}Q(z) = r(z) + \gamma \widehat{P}V(z)$, where $V(x) = \max_{a \in \mathcal{A}}(Q(x, a))$ for all $x \in \mathcal{X}$.

[6]For every real number $u$, $\lceil u \rceil$ is defined as the smallest integer number not less than $u$.

[7]The algorithm $\mathfrak{A}$, unlike QVI, does not need to generate a same number of transition samples for every state-action pair and can generate samples in an arbitrarily way.

## 4. Analysis

In this section, we first provide the full proof of the finite-time PAC bound of QVI, reported in Theorem 1, in Subsection 4.1. We then prove Theorem 2, the RL lower bound, in Subsection 4.2. Also, we need to emphasize that we discard some of the proofs of the technical lemmas due to the lack of space. We provide those results in a long version of this paper.

### 4.1. Poof of Theorem 1

We begin by introducing some new notation. Consider the stationary policy $\pi$. We define $\mathbb{V}^\pi(z) \triangleq \mathbb{E}[|\sum_{t \geq 0} \gamma^t r(Z_t) - Q^\pi(z)|^2]$ as the variance of the sum of discounted rewards starting from $z \in \mathcal{Z}$ under the policy $\pi$. Also, define $\sigma^\pi(z) \triangleq \gamma^2 \sum_{y \in \mathcal{Z}} P^\pi(y|z)|Q^\pi(y) - P^\pi Q^\pi(z)|^2$ as the immediate variance at $z \in \mathcal{Z}$, i.e., $\gamma^2 \mathbb{V}_{Y \sim P^\pi(\cdot|z)}[Q^\pi(Y)]$. Also, we shall denote $v^\pi$ and $v^*$ as the immediate variance of the value function $V^\pi$ and $V^*$ defined as $v^\pi(z) \triangleq \gamma^2 \mathbb{V}_{Y \sim P(\cdot|z)}[V^\pi(Y)]$ and $v^*(z) \triangleq \gamma^2 \mathbb{V}_{Y \sim P(\cdot|z)}[V^*(Y)]$, for all $z \in \mathcal{Z}$, respectively. Further, we denote the immediate variance of the action-value function $\widehat{Q}^\pi$, $\widehat{V}^\pi$ and $\widehat{V}^*$ by $\widehat{\sigma}^\pi$, $\widehat{v}^\pi$ and $\widehat{v}^*$, respectively.

We now state our first result which indicates that $Q_k$ is very close to $\widehat{Q}^*$ up to an order of $O(\gamma^k)$. Therefore, to prove bound on $\|Q^* - Q_k\|$, one only needs to bound $\|Q^* - \widehat{Q}^*\|$ in high probability.

**Lemma 1.** *Let Assumption 1 hold and $Q_0(z)$ be in the interval $[0, \beta]$ for all $z \in \mathcal{Z}$. Then we have*
$$\|Q_k - \widehat{Q}^*\| \leq \gamma^k \beta.$$

In the rest of this subsection, we focus on proving a high probability bound on $\|Q^* - \widehat{Q}^*\|$. One can prove a crude bound of $\tilde{O}(\beta^2/\sqrt{n})$ on $\|Q^* - \widehat{Q}^*\|$ by first proving that $\|Q^* - \widehat{Q}^*\| \leq \beta \|(P - \widehat{P})V^*\|$ and then using the Hoeffding's tail inequality (Cesa-Bianchi & Lugosi, 2006, appendix, pg. 359) to bound the random variable $\|(P - \widehat{P})V^*\|$ in high probability. Here, we follow a different and more subtle approach to bound $\|Q^* - \widehat{Q}^*\|$, which leads to a tight bound of $\tilde{O}(\beta^{1.5}/\sqrt{n})$: **(i)** We prove in Lemma 2 component-wise upper and lower bounds on the error $Q^* - \widehat{Q}^*$ which are expressed in terms of $(I - \gamma \widehat{P}^{\pi^*})^{-1}[P - \widehat{P}]V^*$ and $(I - \gamma \widehat{P}^{\widehat{\pi}^*})^{-1}[P - \widehat{P}]V^*$, respectively. **(ii)** We make use of the sharp result of Bernstein's inequality to bound $[P - \widehat{P}]V^*$ in terms of the squared root of the variance of $V^*$ in high probability. **(iii)** We prove the key result of this subsection (Lemma 5) which shows that the variance of the sum of discounted rewards satisfies a Bellman-like recursion, in which the instant reward $r(z)$ is replaced by $\sigma^\pi(z)$. Based on this result we prove an upper-bound of order $O(\beta^{1.5})$ on $(I-\gamma P^\pi)^{-1}\sqrt{\mathbb{V}(Q^\pi)}$ for any policy $\pi$, which combined with the previous steps leads to the sharp upper bound of $\tilde{O}(\beta^{1.5}/\sqrt{n})$ on $\|Q^*-\widehat{Q}^*\|$.

We proceed by the following lemma which bounds $Q^*-\widehat{Q}^*$ from above and below:

**Lemma 2** (Component-wise bounds on $Q^*-\widehat{Q}^*$).

$$Q^* - \widehat{Q}^* \leq \gamma(I-\gamma\widehat{P}^{\pi^*})^{-1}[P-\widehat{P}]V^*, \quad (1)$$

$$Q^* - \widehat{Q}^* \geq \gamma(I-\gamma\widehat{P}^{\widehat{\pi}^*})^{-1}[P-\widehat{P}]V^*. \quad (2)$$

We now concentrate on bounding the RHS (right hand sides) of (1) and (2). We first state the following technical result which relates $v^*$ to $\widehat{\sigma}^{\pi^*}$ and $\widehat{\sigma}^*$.

**Lemma 3.** *Let Assumption 1 hold and $0 < \delta < 1$. Then, w.p. at least $1 - \delta$:*

$$v^* \leq \widehat{\sigma}^{\pi^*} + b_v \mathbf{1}, \quad (3)$$

$$v^* \leq \widehat{\sigma}^* + b_v \mathbf{1}, \quad (4)$$

*where $b_v$ is defined as*

$$b_v \triangleq \sqrt{\frac{18\gamma^4\beta^4 \log\frac{3N}{\delta}}{n}} + \frac{4\gamma^2\beta^4 \log\frac{3N}{\delta}}{n},$$

*and $\mathbf{1}$ is a function which assigns 1 to all $z \in \mathcal{Z}$.*

Based on Lemma 3 we prove the following sharp bound on $\gamma(P-\widehat{P})V^*$, for which we also rely on Bernstein's inequality (Cesa-Bianchi & Lugosi, 2006, appendix, pg. 361).

**Lemma 4.** *Let Assumption 1 hold and $0 < \delta < 1$. Define $c_{pv} \triangleq 2\log(2N/\delta)$ and $b_{pv}$ as:*

$$b_{pv} \triangleq \left(\frac{6(\gamma\beta)^{4/3}\log\frac{6N}{\delta}}{n}\right)^{3/4} + \frac{5\gamma\beta^2 \log\frac{6N}{\delta}}{n}.$$

*Then w.p. $1-\delta$ we have*

$$\gamma(P-\widehat{P})V^* \leq \sqrt{\frac{c_{pv}\widehat{\sigma}^{\pi^*}}{n}} + b_{pv}\mathbf{1}, \quad (5)$$

$$\gamma(P-\widehat{P})V^* \geq -\sqrt{\frac{c_{pv}\widehat{\sigma}^*}{n}} - b_{pv}\mathbf{1}. \quad (6)$$

*Proof.* For all $z \in \mathcal{Z}$ and all $0 < \delta < 1$, Bernstein's inequality implies that w.p. at least $1-\delta$:

$$(P-\widehat{P})V^*(z) \leq \sqrt{\frac{2v^*(z)\log\frac{1}{\delta}}{\gamma^2 n}} + \frac{2\beta\log\frac{1}{\delta}}{3n},$$

$$(P-\widehat{P})V^*(z) \geq -\sqrt{\frac{2v^*(z)\log\frac{1}{\delta}}{\gamma^2 n}} - \frac{2\beta\log\frac{1}{\delta}}{3n}.$$

We deduce (using a union bound):

$$\gamma(P-\widehat{P})V^* \leq \sqrt{c'_{pv}\frac{v^*}{n}} + b'_{pv}\mathbf{1}, \quad (7)$$

$$\gamma(P-\widehat{P})V^* \geq -\sqrt{c'_{pv}\frac{v^*}{n}} - b'_{pv}\mathbf{1}, \quad (8)$$

where $c'_{pv} \triangleq 2\log(N/\delta)$ and $b'_{pv} \triangleq 2\gamma\beta\log(N/\delta)/3n$. The result then follows by plugging (3) and (4) into (7) and (8), respectively, and then taking a union bound. □

We now state the key lemma of this section which shows that for any policy $\pi$ the variance $\mathbb{V}^\pi$ satisfies the following Bellman-like recursion. Later, we use this result, in Lemma 6, to bound $(I-\gamma P^\pi)^{-1}\sigma^\pi$.

**Lemma 5.** $\mathbb{V}^\pi$ *satisfies the Bellman equation*

$$\mathbb{V}^\pi = \sigma^\pi + \gamma^2 P^\pi \mathbb{V}^\pi.$$

*Proof.* For all $z \in \mathcal{Z}$ we have

$$\mathbb{V}^\pi(z) = \mathbb{E}\left[\left|\sum_{t\geq 0}\gamma^t r(Z_t) - Q^\pi(z)\right|^2\right]$$

$$= \mathbb{E}_{Z_1\sim P^\pi(\cdot|z)}\mathbb{E}\left[\left|\sum_{t\geq 1}\gamma^t r(Z_t) - \gamma Q^\pi(Z_1)\right.\right.$$

$$\left.\left.-(Q^\pi(z)-r(z)-\gamma Q^\pi(Z_1))\right|^2\right]$$

$$= \gamma^2\mathbb{E}_{Z_1\sim P^\pi(\cdot|z)}\mathbb{E}\left[\left|\sum_{t\geq 1}\gamma^{t-1}r(Z_t) - Q^\pi(Z_1)\right|^2\right]$$

$$-2\mathbb{E}_{Z_1\sim P^\pi(\cdot|z)}\left[(Q^\pi(z)-r(z)-\gamma Q^\pi(Z_1))\right.$$

$$\left.\times \mathbb{E}\left(\sum_{t\geq 1}\gamma^t r(Z_t) - \gamma Q^\pi(Z_1)\Big|Z_1\right)\right]$$

$$+\mathbb{E}_{Z_1\sim P^\pi(\cdot|z)}(|Q^\pi(z)-r(z)-\gamma Q^\pi(Z_1)|^2)$$

$$= \gamma^2\mathbb{E}_{Z_1\sim P^\pi(\cdot|z)}\mathbb{E}\left[\left|\sum_{t\geq 1}\gamma^{t-1}r(Z_t) - Q^\pi(Z_1)\right|^2\right]$$

$$+\gamma^2\mathbb{V}_{Z_1\sim P^\pi(\cdot|z)}(Q^\pi(Z_1))$$

$$= \gamma^2\sum_{y\in\mathcal{Z}}P^\pi(y|z)\mathbb{V}^\pi(y) + \sigma^\pi(z),$$

in which we rely on $\mathbb{E}(\sum_{t\geq 1}\gamma^t r(Z_t) - \gamma Q^\pi(Z_1)|Z_1) = 0$. □

Based on Lemma 5, one can prove the following result on the immediate variance.

**Lemma 6.**

$$\|(I-\gamma^2 P^\pi)^{-1}\sigma^\pi\| \leq \beta^2, \quad (9)$$

$$\|(I-\gamma P^\pi)^{-1}\sqrt{\sigma^\pi}\| \leq 2\log(2)\beta^{1.5}. \quad (10)$$

*Proof.* The first inequality follows from Lemma 5 by solving (5) in terms of $\mathbb{V}^\pi$ and taking the sup-norm over both sides of the resulted equation. In the case of Eq.(10) we have [8]

$$\|(I - \gamma P^\pi)^{-1}\sqrt{\sigma^\pi}\| = \left\|\sum_{k \geq 0}(\gamma P^\pi)^k \sqrt{\sigma^\pi}\right\|$$

$$= \left\|\sum_{l \geq 0}(\gamma P^\pi)^{tl} \sum_{j=0}^{t-1}(\gamma P^\pi)^j \sqrt{\sigma^\pi}\right\|$$

$$\leq \sum_{l \geq 0}(\gamma^t)^l \left\|\sum_{j=0}^{t-1}(\gamma P^\pi)^j \sqrt{\sigma^\pi}\right\|$$

$$= \frac{1}{1-\gamma^t}\left\|\sum_{j=0}^{t-1}(\gamma P^\pi)^j \sqrt{\sigma^\pi}\right\|, \quad (11)$$

in which we write $k = tl + j$ with $t$ is a positive integer. We now prove a bound on $\|\sum_{j=0}^{t-1}(\gamma P^\pi)^j \sqrt{\sigma^\pi}\|$ by making use of Jensen's inequality as well as Cauchy-Schwarz inequality:

$$\left\|\sum_{j=0}^{t-1}(\gamma P^\pi)^j \sqrt{\sigma^\pi}\right\| \leq \left\|\sum_{j=0}^{t-1}\gamma^j \sqrt{(P^\pi)^j \sigma^\pi}\right\|$$

$$\leq \sqrt{t}\left\|\sqrt{\sum_{j=0}^{t-1}(\gamma^2 P^\pi)^j \sigma^\pi}\right\| \quad (12)$$

$$\leq \sqrt{t}\left\|\sqrt{(I - \gamma^2 P^\pi)^{-1}\sigma^\pi}\right\|$$

$$\leq \beta\sqrt{t},$$

where in the last step we rely on (9). The result then follows by plugging (12) into (11) and optimizing the bound in terms of $t$ to achieve the best dependency on $\beta$. □

Now we make use of Lemma 6 and Lemma 4 to bound $\|Q^* - \widehat{Q}^*\|$ in high probability.

**Lemma 7.** *Let Assumption 1 hold. Then, for any $0 < \delta < 1$:*

$$\|Q^* - \widehat{Q}^*\| \leq \varepsilon',$$

*w.p. $1 - \delta$, where $\varepsilon'$ is defined as:*

$$\varepsilon' \triangleq \sqrt{\frac{17\beta^3 \log \frac{4N}{\delta}}{n}} + \left(\frac{6(\gamma\beta^2)^{4/3} \log \frac{12N}{\delta}}{n}\right)^{3/4}$$

$$+ \frac{5\gamma\beta^3 \log \frac{12N}{\delta}}{n}. \quad (13)$$

---
[8] For any real-valued function $f$, $\sqrt{f}$ is defined as a component wise squared-root operator on $f$.

*Proof.* By incorporating the result of Lemma 4 and Lemma 6 into Lemma 2, we deduce that:

$$Q^* - \widehat{Q}^* \leq b\mathbf{1},$$
$$Q^* - \widehat{Q}^* \geq -b\mathbf{1},$$

w.p. $1 - \delta$. The scalar $b$ is given by:

$$b \triangleq \sqrt{\frac{17\beta^3 \log \frac{2N}{\delta}}{n}} + \left(\frac{6(\gamma\beta^2)^{4/3} \log \frac{6N}{\delta}}{n}\right)^{3/4}$$

$$+ \frac{5\gamma\beta^3 \log \frac{6N}{\delta}}{n}.$$

The result then follows by combining these two bounds and taking the $\ell_\infty$ norm. □

*Proof of Theorem 1.* We combine the proof of Lemma 7 and Lemma 1 in order to bound $Q^* - Q_k$ in high probability. We then solve the resulted bound w.r.t. $n$ and $k$.[9] □

### 4.2. Proof of the Lower-bound

In this section, we provide the proof of Theorem 2. In our analysis, we rely on the likelihood-ratio method, which has been previously used to prove a lower bound for multi-armed bandits (Mannor & Tsitsiklis, 2004), and extend this approach to RL and MDPs. We begin by defining a class of MDPs for which the proposed lower bound will be obtained (see Figure 1). We define the class of MDPs $\mathbb{M}$ as the set of all MDPs with the state-action space of cardinality $N = 3KL$, where $K$ and $L$ are positive integers. Also, we assume that for all $M \in \mathbb{M}$, the state space $\mathcal{X}$ consists of three smaller sets $\mathcal{S}$, $\mathcal{Y}_1$ and $\mathcal{Y}_2$. The set $\mathcal{S}$ includes $K$ states, each of those states corresponds with the set of actions $\mathcal{A} = \{a_1, a_2, \ldots, a_L\}$, whereas the states in $\mathcal{Y}^1$ and $\mathcal{Y}^2$ are single-action states. By taking the action $a \in \mathcal{A}$ from every state $x \in \mathcal{S}$, we move to the next state $y(z) \in \mathcal{Y}^1$ with the probability 1, where $z = (x, a)$. The transition probability from $\mathcal{Y}^1$ is characterized by the transition probability $p_M$ from every $y(z) \in \mathcal{Y}^1$ to itself and with the probability $1 - p_M$ to the corresponding $y(z) \in \mathcal{Y}^2$.[10] Further, for all $M \in \mathbb{M}$, $\mathcal{Y}^2$ consists of only absorbing states, i.e., for all $y \in \mathcal{Y}^2$, $P(x|x) = 1$. The instant reward $r$ is set to 1 for every state in $\mathcal{Y}^1$ and 0 elsewhere. For this class of MDPs, the optimal action-value function $Q^*$ can be solved in close form from the Bellman equation:

$$Q^*(z) = \gamma V^*(y(z)) = \frac{\gamma}{1 - \gamma p_M}, \quad \forall z \in \mathcal{S} \times \mathcal{A},$$

---
[9] Note that the total number of samples is then computed by $T = Nn$.
[10] Every state $y \in \mathcal{Y}^2$ is only connected to one state in $\mathcal{Y}^1$ and $\mathcal{S}$, i.e., there is no overlapping path in the MDP.

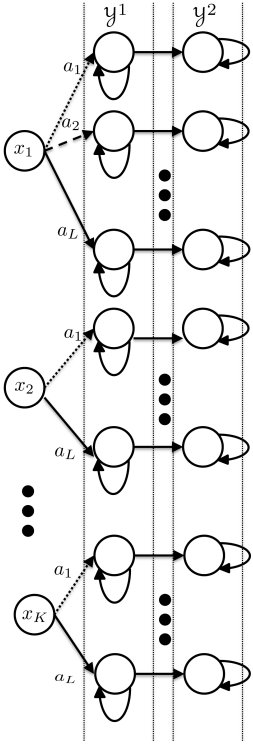

Figure 1. The class of MDPs considered in the proof of Theorem 2. Nodes represent states and arrows show transitions between the states (see the text for details).

In the rest of the proof, we concentrate on proving the lower bound for $\|Q^* - Q_T^{\mathfrak{A}}\|$ for all $z \in \mathcal{S} \times \mathcal{A}$. Now, let us consider a set of two MDPs $\mathbb{M}^* = \{M_0, M_1\}$ in $\mathbb{M}$ with the transition probabilities

$$p_M = \begin{cases} p & M = M_0, \\ p + \alpha & M = M_1, \end{cases}$$

where $\alpha$ and $p$ are some positive numbers such that $0 < p < p + \alpha \leq 1$, which will be quantified later in this section. We assume that the discount factor $\gamma$ is bounded from below by some positive constant $\gamma_0$ for the class of MDP $M^*$. We denote by $\mathbb{E}_m$ ad $\mathbb{P}_m$ the expectation and the probability under the model $M_m$ in the rest of this section.

We follow the following steps in the proof: **(i)** we prove a lower bound on the sample-complexity of learning the value function for every state $y \in \mathcal{Y}$ on the class of MDP $\mathbb{M}^*$ **(ii)** we then make use of the fact that the estimates of $Q^*(z)$ for different $z \in \mathcal{S} \times \mathcal{A}$ are independent of each others to combine these bounds and prove the tight result of Theorem 2.

We now introduce some new notation: Let $Q_t^{\mathfrak{A}}(z)$ be an empirical estimate of the action-value function $Q^*(z)$ by the RL algorithm $\mathfrak{A}$ using $t > 0$ transition samples from the state $y(z) \in \mathcal{Y}^1$ for $z \in \mathcal{X} \times \mathcal{A}$. We define the event $\mathcal{E}_1(z) \triangleq \{|Q_0^*(z) - Q_t^{\mathfrak{A}}(z)| \leq \varepsilon\}$ for all $z \in \mathcal{S} \times \mathcal{A}$, where $Q_0^* \triangleq \gamma/(1-\gamma p)$ is the optimal action-value function for all $z \in \mathcal{S} \times \mathcal{A}$ under the MDP $M_0$. We then define $k \triangleq r_1 + r_2 + \cdots + r_t$ as the sum of rewards of making $t$ transitions from $y(z) \in \mathcal{Y}^1$. We also introduce the event $\mathcal{E}_2(z)$, for all $z \in \mathcal{S} \times \mathcal{A}$ as:

$$\mathcal{E}_2(z) \triangleq \left\{ pt - k \leq \sqrt{2p(1-p)t \log \frac{c_2'}{2\theta}} \right\}.$$

Further, we define $\mathcal{E}(z) \triangleq \mathcal{E}_1(z) \cap \mathcal{E}_2(z)$. We then state the following lemmas required for our analysis.

We begin our analysis of the lower bound by the following lemma:

**Lemma 8.** *Define $\theta \triangleq \exp\left(-c_1' \alpha^2 t/(p(1-p))\right)$. Then, for every RL algorithm $\mathfrak{A}$, there exists an MDP $M_m \in \mathbb{M}^*$ and constants $c_1' > 0$ and $c_2' > 0$ such that*

$$\mathbb{P}_m(|Q^*(z) - Q_t^{\mathfrak{A}}(z)|) > \varepsilon) > \frac{\theta}{c_2'}, \qquad (14)$$

*by the choice of $\alpha = 2(1 - \gamma p)^2 \varepsilon/(\gamma^2)$.*

*Proof.* To prove this result we make use of a contradiction argument, i.e., we assume that there exists an algorithm $\mathfrak{A}$ for which:

$$\mathbb{P}_m((|Q^*(z) - Q_t^{\mathfrak{A}}(z)|) > \varepsilon) \leq \frac{\theta}{c_2'}, \quad \text{or}$$
$$\mathbb{P}_m((|Q^*(z) - Q_t^{\mathfrak{A}}(z)|) \leq \varepsilon) \geq 1 - \frac{\theta}{c_2'}, \qquad (15)$$

for all $M_m \in \mathbb{M}^*$ and show that this assumption leads to a contradiction. We now state the following technical lemmas required for the proof:

**Lemma 9.** *For all $p > \frac{1}{2}$:*

$$\mathbb{P}_0(\mathcal{E}_2(z)) > 1 - \frac{2\theta}{c_2'}.$$

Now, by the assumption that $\mathbb{P}_m(|Q^*(z) - Q_t^{\mathfrak{A}}(z)|) > \varepsilon) \leq \theta/c_2'$ for all $M_m \in \mathbb{M}^*$, we have $\mathbb{P}_0(\mathcal{E}_1(z)) \geq 1 - \theta/c_2' \geq 1 - 1/c_2'$. This combined with Lemma 9 and with the choice of $c_2' = 6$ implies that $\mathbb{P}_0(\mathcal{E}(z)) > 1/2$, for all $z \in \mathcal{S} \times \mathcal{A}$.

**Lemma 10.** *Let $\varepsilon \leq \frac{1-p}{4\gamma^2(1-\gamma p)^2}$. Then, for all $z \in \mathcal{S} \times \mathcal{A}$: $\mathbb{P}_1(\mathcal{E}_1(z)) > \theta/c_2'$.*

Now by the choice of $\alpha = 2(1 - \gamma p)^2 \varepsilon/(\gamma^2)$, we have that $Q_1^*(z) - Q_0^*(z) = \frac{\gamma}{1-\gamma(p+\alpha)} - \frac{\gamma}{1-\gamma p} > 2\varepsilon$, thus $Q_0^*(z) + \varepsilon < Q_1^*(z) - \varepsilon$. In words, the random event $\{|Q_0^*(z) - Q_t^{\mathfrak{A}}(z)| \leq \varepsilon\}$ does not overlap with the event $\{|Q_1^*(z) - Q_t^{\mathfrak{A}}(z)| \leq \varepsilon\}$.

Now let us return to the assumption of Eq. (15), which states that for all $M_m \in \mathbb{M}^*$, $\mathbb{P}_m(|Q^*(z) - Q_t^{\mathfrak{A}}(z)|) \leq \varepsilon) \geq 1 - \theta/c_2'$ under Algorithm $\mathfrak{A}$. Based on Lemma 10 we have $\mathbb{P}_1(|Q_0^*(z) - Q_t^{\mathfrak{A}}(z)| \leq \varepsilon) > \theta/c_2'$. This combined with the fact that $\{|Q_0^*(y) - Q_t^{\mathfrak{A}}(z)|\}$ and $\{|Q_1^*(z) - Q_t^{\mathfrak{A}}(z)|\}$ do not overlap implies that $\mathbb{P}_1(|Q^*(z) - Q_t^{\mathfrak{A}}(z)|) \leq \varepsilon) \leq 1 - \theta/c_2'$, which violates the assumption of Eq. (15). The contradiction between the result of Lemma 10 and the assumption which leads to this result proves the lower bound of Eq. (14). □

Now by the choice of $p = \frac{4\gamma - 1}{3\gamma}$ and $c_1 = 8100$, we have that for every $\varepsilon \in (0, 3]$ and for all $0.4 = \gamma_0 \leq \gamma < 1$ there exists an MDP $M_m \in \mathbb{M}^*$ such that

$$\mathbb{P}_m(|Q^*(z) - Q_{T_z}^{\mathfrak{A}}(z)|) > \varepsilon) > \frac{1}{c_2'} e^{\frac{-c_1 T_z \varepsilon^2}{6\beta^3}},$$

This result implies that for any state-action $z \in \mathcal{S} \times \mathcal{A}$, the probability of making an estimation error of $\varepsilon$ is at least $\delta$ on $M_0$ or $M_1$ whenever the number of transition samples $T_z$ from $z \in \mathcal{Z}$ is less that $\xi(\varepsilon, \delta) \triangleq \frac{6\beta^3}{c_1 \varepsilon^2} \log \frac{1}{c_2' \delta}$. We now extend this result to the whole state-action space $\mathcal{S} \times \mathcal{A}$.

**Lemma 11.** *Assume that for every algorithm $\mathfrak{A}$, for every state-action $z \in \mathcal{S} \times \mathcal{A}$ we have*[11]

$$\mathbb{P}_m(|Q^*(z) - Q_{T_z}^{\mathfrak{A}}(z)| > \varepsilon | T_z = t_z) > \delta, \quad (16)$$

*Then for any $\delta' \in (0, 1/2)$, for any algorithm $\mathfrak{A}$ using a total number of transition samples less than $T = \frac{N}{6}\xi(\varepsilon, \frac{12\delta'}{N})$, there exists an MDP $M_m \in \mathbb{M}^*$ such that*

$$\mathbb{P}_m(\|Q^* - Q_T^{\mathfrak{A}}\| > \varepsilon) > \delta', \quad (17)$$

*where $Q_T^{\mathfrak{A}}$ denotes the empirical estimate of the optimal action-value function $Q^*$ by $\mathfrak{A}$ using $T$ transition samples.*

*Proof.* First note that if the total number of observed transitions is less than $KL/2\xi(\varepsilon, \delta) = (N/6)\xi(\varepsilon, \delta)$, then there exists at least $KL/2 = N/6$ state-action pairs that are sampled at most $\xi(\varepsilon, \delta)$ times. Indeed, if this was not the case, then the total number of transitions would be strictly larger than $N/6\xi(\varepsilon, \delta)$, which implies a contradiction). Now let us denote those states as $z_{(1)}, \ldots, z_{(N/6)}$.

We consider the specific class of MDPs described in Figure 1. In order to prove that (17) holds for any algorithm, it is sufficient to prove it for the class of algorithms that return an estimate $Q_{T_z}^{\mathfrak{A}}(z)$ for each state-action $z$ based on the transition samples observed from $z$ only (indeed, since the samples from $z$ and $z'$ are independent, the samples collected from $z'$ do not bring more information about $Q^*(z)$ than the information brought by the samples collected from $z$). Thus, by defining $\mathcal{Q}(z) \triangleq \{|Q^*(z) - Q_{T_z}^{\mathfrak{A}}(z)| > \varepsilon\}$, we have that for such algorithms, the events $\mathcal{Q}(z)$ and $\mathcal{Q}(z')$ are conditionally independent given $T_z$ and $T_{z'}$. Thus, there exists an MDP $M_m \in \mathbb{M}^*$ such that:

$$\mathbb{P}_m\Big(\{\mathcal{Q}(z_{(i)})^c\}_{1 \leq i \leq \frac{N}{6}} \cap \{T_{z(i)} \leq \xi(\varepsilon, \delta)\}_{1 \leq i \leq \frac{N}{6}}\Big)$$

$$= \sum_{t_1=0}^{\xi(\varepsilon,\delta)} \cdots \sum_{t_{N/6}=0}^{\xi(\varepsilon,\delta)} \mathbb{P}_m\Big(\{T_{z(i)} = t_i\}_{1 \leq i \leq \frac{N}{6}}\Big)$$

$$\mathbb{P}_m\Big(\{\mathcal{Q}(z_{(i)})^c\}_{1 \leq i \leq N/6} \cap \{T_{z(i)} = t_i\}_{1 \leq i \leq \frac{N}{6}}\Big)$$

$$= \sum_{t_1=0}^{\xi(\varepsilon,\delta)} \cdots \sum_{t_{N/6}=0}^{\xi(\varepsilon,\delta)} \mathbb{P}_m\Big(\{T_{z(i)} = t_i\}_{1 \leq i \leq \frac{N}{6}}\Big)$$

$$\prod_{1 \leq i \leq N/6} \mathbb{P}_m\Big(\mathcal{Q}(z_{(i)})^c \cap T_{z(i)} = t_i\Big)$$

$$\leq \sum_{t_1=0}^{\xi(\varepsilon,\delta)} \cdots \sum_{t_{N/6}=0}^{\xi(\varepsilon,\delta)} \mathbb{P}_m\Big(\{T_{z(i)} = t_i\}_{1 \leq i \leq \frac{N}{6}}\Big)(1 - \delta)^{\frac{N}{6}},$$

from Eq. (16), thus

$$\mathbb{P}_m\Big(\{\mathcal{Q}(z_{(i)})^c\}_{1 \leq i \leq N/6}\big|\{T_{z(i)} \leq \xi(\varepsilon, \delta)\}_{1 \leq i \leq N/6}\Big)$$
$$\leq (1 - \delta)^{N/6}.$$

---
[11]Note that we allow $T_z$ to be random.

We finally deduce that if the total number of transition samples is less than $\frac{N}{6}\xi(\varepsilon, \delta)$, then

$$\mathbb{P}_m(\|Q^* - Q_T^{\mathfrak{A}}\| > \varepsilon) \geq \mathbb{P}_m\Big(\bigcup_{z \in \mathcal{S} \times \mathcal{A}} \mathcal{Q}(z)\Big)$$

$$\geq 1 - \mathbb{P}_m\Big(\{\mathcal{Q}(z_{(i)})^c\}_{1 \leq i \leq \frac{N}{6}}\big|\{T_{z(i)} \leq \xi(\varepsilon, \delta)\}_{1 \leq i \leq \frac{N}{6}}\Big)$$

$$\geq 1 - (1 - \delta)^{\frac{N}{6}} \geq \frac{\delta N}{12},$$

whenever $\frac{\delta N}{6} \leq 1$. Setting $\delta' = \frac{\delta N}{12}$, we obtain the desired result. □

Lemma 11 implies that if the total number of samples $T$ is less than $\beta^3 N/(c_1 \varepsilon^2) \log(N/(c_2 \delta))$, with the choice of $c_1 = 8100$ and $c_2 = 72$, then the probability of $\|Q^* - Q_T^{\mathfrak{A}}\| \leq \varepsilon$ is at maximum $1 - \delta$ on either $M_0$ or $M_1$. This is equivalent to the statement that for every RL algorithm $\mathfrak{A}$ to be $(\varepsilon, \delta, T)$-correct on the set $\mathbb{M}^*$, and subsequently on the class of MDPs $\mathbb{M}$, the total number of transitions $T$ needs to satisfy the inequality $T > \beta^3 N/(c_1 \varepsilon^2) \log(N/(c_2 \delta))$, which concludes the proof of Theorem 2.

## 5. Conclusion and Future Works

In this paper, we have presented the first minimax bound on the sample complexity of estimating the optimal action-value function in discounted reward MDPs. We have proven that the model-based Q-value iteration algorithm (QVI) is an optimal learning algorithm since it minimizes the dependencies on $1/\varepsilon$, $N$, $\delta$ and $\beta$. Also, our results have significantly improved on the state-of-the-art in terms of dependency on $\beta$. Overall, we conclude that QVI is an efficient RL algorithm which completely closes the gap between the lower and upper bound of the sample complexity of RL in the presence of a generative model of the MDP.

In this work, we are only interested in the estimation of the optimal action-value function and not the problem of exploration. Therefore, we did not compare our results with the-state-of-the-art of PAC-MDP (Strehl et al., 2009; Szita & Szepesvári, 2010) and upper-confidence bound based algorithms (Bartlett & Tewari, 2009; Jaksch et al., 2010), in which the choice of the exploration policy has an influence on the behavior of the learning algorithm. However, we believe that it would be possible to improve on the state-of-the-art in PAC-MDP, based on the results of this paper. This is mainly due to the fact that most PAC-MDP algorithms rely on an extended variant of model-based Q-value iteration to estimate the action-value function, but they use the naive result of Hoeffding's inequality for concentration of measure which leads to non-tight sample complexity results. One can improve on those results, in terms of dependency on $\beta$, using the improved analysis of this paper which makes use of the sharp result of Bernstein's inequality as opposed to the Hoeffding's inequality in the previous works. Also, we believe that the existing lower bound on the *sample complexity of exploration* of any reinforcement learning algorithm (Strehl et al., 2009) can be significantly improved in terms of dependency on $\beta$ using the new "hard" class of MDPs presented in this paper.

### Acknowledgments

The authors appreciate supports from the European Community's Seventh Framework Programme (FP7/2007-2013) under grant agreement n° 231495. We also thank Bruno Scherrer for useful discussion and the anonymous reviewers for their valuable comments.